\documentclass{article}
\usepackage{ijcai25}

\usepackage{times}
\usepackage{soul}
\usepackage{url}
\usepackage[utf8]{inputenc}
\usepackage[small]{caption}
\usepackage{graphicx}
\usepackage{amsmath}
\usepackage{amsthm}
\usepackage{booktabs}
\usepackage{algorithm}
\usepackage{algorithmic}
\usepackage[switch]{lineno}
\usepackage{xcolor} 
\usepackage{tcolorbox}
\usepackage{pdflscape}  
\usepackage{lscape} 
\usepackage{comment}
\usepackage{float}
\usepackage{subcaption}
\usepackage{multirow}

\usepackage{adjustbox}

\title{Perspectives in Play:\\A Multi-Perspective Approach for More  Inclusive NLP Systems}

\author{
Benedetta Muscato$^{1,2}$ \thanks{Corresponding Author: \text{benedetta.muscato@sns.it}}
\and
Lucia Passaro$^2$\and
Gizem Gezici$^1$\And
Fosca Giannotti$^1$\\
\affiliations
$^1$Scuola Normale Superiore, Italy\\
$^2$University of Pisa, Italy\\
\emails
\{benedetta.muscato, gizem.gezici, fosca.giannotti\}@sns.it,
lucia.passaro@unipi.it
}

\begin{document}

\maketitle

\begin{abstract}
    In the realm of Natural Language Processing (NLP), common approaches for handling human disagreement consist of aggregating annotators' viewpoints to establish a single ground truth. However, prior studies show that disregarding individual opinions can lead to the side-effect of under-representing minority perspectives, especially in subjective tasks, where annotators may systematically disagree because of their preferences. Recognizing that labels reflect the diverse backgrounds, life experiences, and values of individuals, this study proposes a new multi-perspective approach using soft labels to encourage the development of the next generation of perspective-aware models—more  inclusive and pluralistic. We conduct an extensive analysis across diverse subjective text classification tasks including hate speech, irony, abusive language, and stance detection, to highlight the importance of capturing human disagreements, often overlooked by traditional aggregation methods.
    Results show that the multi-perspective approach not only better approximates human label distributions, as measured by Jensen-Shannon Divergence (JSD), but also achieves superior classification performance (higher F1-scores), outperforming traditional approaches.
    However, our approach exhibits lower confidence in tasks like irony and stance detection, likely due to the inherent subjectivity present in the texts. Lastly, leveraging Explainable AI (XAI), we explore model uncertainty and uncover meaningful insights into model predictions. All implementation details are available at~\texttt{\url{https://github.com/bmuscato/IJCAI_25_MultiPerspective}}.
\end{abstract}

\noindent \textbf{Warning:} \textit{{This paper contains examples that may be offensive or upsetting.}}

\section{Introduction}

Recent studies in NLP have highlighted the significance of annotator disagreement, reframing it as a form of plausible and rich~\emph{human label variation} (HLV), rather than noise ~\cite{plank2022problem}.
Research has shown that Large Language Models (LLMs) can be biased towards certain viewpoints~\cite{santurkar2023whose}, underscoring disparities which may affect underrepresented communities and potentially reinforce existing societal prejudices~\cite{aoyagui2024exploring}. 
As LLMs continue to advance, aligning them with human subjectivity—encompassing diverse perspectives, individual values, and pluralistic principles\footnote{A system is considered pluralistic if it is designed to accommodate a broad range of human values and viewpoints~\cite{sorensen2024roadmap}.}—becomes increasingly crucial~\cite{kirk:benefits,wang:aligning}.

To address this challenge,~\emph{Perspectivism}~\cite{basile2021toward}, an emerging multidisciplinary approach in NLP, leverages disaggregated datasets\footnote{
In human-labeled datasets, disaggregated labels keep each annotator's input, avoiding to collapse them into one label through majority voting.} that capture human disagreements. By doing so, it amplifies diverse voices while also considering sociodemographic factors and ethical concerns.
This new paradigm differs from traditional methods that often rely on a single ground-truth derived from vote aggregation approaches such as majority voting. Instead, perspectivist approach aims to preserve all annotations and embraces diverse opinions through learning from human disagreement~\cite{uma2021learning}, thus avoiding the suppression of minority perspectives. 
To follow the perspectivist paradigm, we present a comprehensive overview of experiments conducted between established baselines from the literature \cite{davani2022dealing}- predicting aggregated labels - and our approach, which models human disagreement in a nuanced, fuzzy manner without assuming the existence of a singular ground truth. To this end, we investigate whether the multi-perspective methodology, through incorporating diverse human opinions, can enhance i) overall model performance, and ii) model confidence. As we are interested in exploring different subjective tasks, we introduce a new version of an existing dataset from \cite{gezici2021evaluation}, which provides a summarized version of the original documents on controversial topics, designed for the stance detection task. To ensure the robustness of our analysis, we also perform ablation studies to test whether model predictions align with human disagreement using eXplainable AI (XAI) techniques, specifically post-hoc feature attribution methods, to enhance the transparency of the proposed models.

\section{Related Work}
In this section, we lay the foundation for our analysis by combining insights on preserving human disagreement in the dataset design process and reviewing the model learning approaches adopted in the literature so far.
\paragraph{Perspective-aware Datasets} 

Disagreement in human labeling, now recognized as \emph{subjectivity} driven by HLV, affects all stages of the NLP pipeline, from data preparation to evaluation. This variability arises from different annotator perceptions, particularly in subjective tasks like hate and emotion detection.
As a result, recent studies have embraced a perspectivist approach~\cite{basile:2020s,leonardelli:semeval,sandri:don,fleisig:majority,abercrombie:consistency,muscato2025embracing} to further explore and incorporate human preferences in various subjective NLP tasks. This has led to the creation of publicly available perspectivist disaggregated datasets\footnote{\texttt{\url{https://pdai.info}}} listed in one of the pioneering perspectivist surveys~\cite{frenda2024perspectivist}, emphasizing the importance of recognizing and assessing human disagreement as a \textit{valuable} and \textit{plausible} aspect of data annotation.

\paragraph{Perspective-aware Models}

The use of human disagreement as a learning signal has been shown to be effective in NLP and other areas of AI \cite{uma2021learning,muscato2024multi}. Strategies including training annotator-specific classifiers~\cite{basile:2020s}, adopting multi-task architectures~\cite{davani2022dealing}, or encoding socio-demographic information about the annotators~\cite{beck2024sensitivity} have been proposed to accommodate annotator diverse preferences. A key distinction lies in the usage of hard labels, i.e. aggregated labels, often criticized for oversimplifying complex data, versus soft labels, which can capture a range of possible values and better reflect the ambiguities. Techniques like Knowledge Distillation~\cite{hinton2015distilling}, Label Smoothing~\cite{szegedy2016rethinking} can improve model performances. Moreover, allowing models to learn from soft labels not only enhances robustness and generalization~\cite{collins2022eliciting,pereyra:regularizing}, but also proves particularly effective for subjective tasks such as detecting stereotypes~\cite{schmeisser2024human}, misogyny, and abusiveness~\cite{leonardelli:semeval}.

\begin{figure*}[!t]
    \centering
    \includegraphics[width=0.66\linewidth]{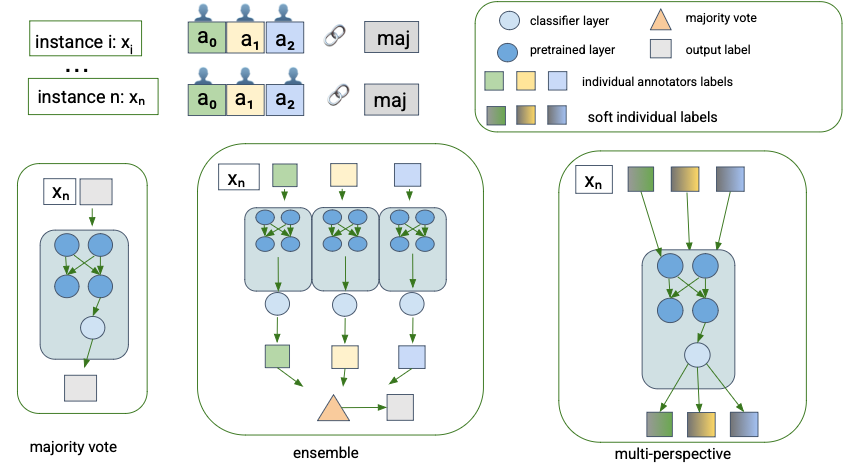}  
 \caption{
 Overview of baselines, left: majority vote, middle: ensemble and right: the multi-perspective approach, which uses soft labels shown as color gradients. Annotator counts are illustrative; actual benchmarks range from 3 to 18 annotators.
 }
 \label{fig:pipeline}
\end{figure*}

\section{Methodology}
\label{sec:metodology}

We define the classification task on an annotated dataset \( D = (X, A, Y) \), where \( X \) represent the set of text instances, \( A \) is the set of annotators, and \( Y \) is the annotation matrix. Each entry \( y_{ij} \) in \( Y \) indicates the annotation provided by annotator \( a_j \in A \) for the text instance \( x_i \in X \). Unlike conventional setups, each \( y_{ij} \) is represented as a probability distribution over the interval \([0, 1]\), reflecting the likelihood that the annotator prefers one class over another. 
Our methodology is two-fold:

\begin{enumerate}
    \item \textbf{Baseline}: Similar to traditional models, our baselines learn from a set of hard labels, which are derived by aggregating all individual annotations of matrix $|Y|$.
    \item \textbf{Multi-Perspective}: The model learns by leveraging the probabilistic version of hard labels  in $|Y|$, i.e., soft labels.
    \end{enumerate}

Figure \ref{fig:pipeline}, inspired by \cite{davani2022dealing}, provides an overview of the baselines and our proposed methodology, which we will elaborate in the following sections.
Specifically, we fine-tune BERT-Large~\cite{devlin2018bert} and RoBERTa-Large~\cite{liu2019roberta}, both encoder-only models, 
particularly suitable for classification tasks, especially when working with relatively small datasets, as in the case of our study.
Additional details regarding the fine-tuning settings are provided in our GitHub repository, linked on the first page.

\subsection{Baseline Models using Hard Labels}
\label{sec:baseline}

In traditional machine learning (ML) settings, the most commonly used label aggregation technique is \emph{majority voting}, where the label that appears most frequently among the annotations provided by annotators is chosen for each data point. Typically, the majority label is represented as a \emph{hard label}, which, in binary classification settings, is encoded as either $0$ or $1$.  
Following \cite{davani2022dealing}, we consider two different baselines: majority vote, a single-task classifier trained to predict
the aggregated label and an ensemble, a multi-annotator classifier whose aim is to predict the aggregated label.
\paragraph{Majority Vote}
The majority vote baseline is a widely used approach in which a single-task classifier is fine-tuned to predict the aggregated label for each instance—in this case, the majority label. 
It is built by adding a fully connected layer to the outputs of the language model (\(h_i\)), followed by the application of a  linear transformation and a softmax function. The softmax generates the probability of the majority label, \( P(\text{maj}(\bar{y}) \mid h) \).
\paragraph{Ensemble} 
The ensemble approach involves predicting the annotations generated by each annotator. During training, each classifier is independently fine-tuned to predict all annotations provided by the \(j\)-th annotator. At test time, the outputs generated by the ensemble are aggregated to determine the majority label, meant to be the final prediction \( P(\text{maj}(\bar{y}) \mid h) \). 

\subsection{Multi-Perspective Model using Soft Labels}
\label{sec:multip}
In the multi-perspective schema, differently from the baseline approach, a majority label is not created, meaning that the multi-perspective model uses disaggregated labels. Moreover, these disaggregated labels which are represented with discrete values are transformed into their continuous counterparts using softmax function and called as~\emph{soft labels} following the approach presented in~\cite{uma2020case}. 
Instead of assigning a single label to each data point, soft labels capture a probability distribution across the possible classes, ensuring that minority perspectives are included rather than being overlooked. This approach is particularly valuable in subjective tasks where annotator preferences may vary significantly~\cite{plank2022problem}.
Since the baseline and multi-perspective approaches design the datasets in a distinct manner- the baseline approach uses hard labels, while the multi-perspective approach uses soft labels- in the multi-perspective settings, the soft loss~\cite{uma2020case} is applied instead of cross-entropy loss. This choice is driven by the need to encode the human label distribution with a more fine-grained representation. The soft loss is defined as:

\begin{equation*}
-\sum_{i=1}^n \sum_{c} p_{\text{hum}}(y_i = c \mid x_i) \log p_\theta(y_i = c \mid x_i)
\end{equation*}

where $p_{\text{hum}}(y \mid x)$ represents the human label distribution (i.e. soft labels) which is obtained by applying the softmax function to the logits produced by the classifier.

\begin{table*}[!t]
    \centering
    \renewcommand{\arraystretch}{0.8}  
        \footnotesize
    \begin{tabular}{lcccccccc}
        \toprule
        Dataset & Class\_0 & Class\_1 & Class\_2 & Class\_3 & Tot.Class & Ann. & \%Disagr. & Subj. Task \\
        \midrule
        GabHate & 25212 & 2452 & -- & -- & 2 & 18 (3.13) & 69\% & Hate speech detection \\
        ConvAbuse & 3378 & 672 & -- & -- & 2 & 8 (3.24) & 46\% & Abusive lang. detection \\
        EPIC & 2228 & 761 & -- & -- & 2 & 5 (4.72) & 66\% &  Irony detection \\
        StanceDetection & 223 & 199 & 298 & 177 & 4 & 3 (3.0) & 73\% & Stance detection \\
        \bottomrule
    \end{tabular}
    \caption{
    Overview of majority label distribution, class count, annotator stats, and disagreement rate—highlighting subjectivity and excluded minority views from majority voting.
    }
    \label{tab:distribution}
\end{table*}

\section{Experimental Setup}
\label{sec:experiments}
This section details the technical aspects of the conducted experiments to evaluate our Multi-Perspective approach. We first provide an overview of the benchmark datasets used, followed by the introduction of our custom dataset for stance detection and its summarization process. Finally, we describe the fine-tuning setup employed in the experiments. Our code, along with supplementary materials - including document summarization, fine-tuning details, and XAI visualizations - is available at the referenced Github page.

\subsection{Datasets}
\label{originaldataset}
In this study, we conduct experiments on datasets related to four distinct subjective tasks: Gab Hate \cite{kennedy2018gab} for hate speech detection, ConvAbuse \cite{cercas-curry-etal-2021-convabuse} for abusive language detection,  EPIC \cite{frenda2023epic} for irony detection and finally our Stance Detection dataset \cite{gezici2021evaluation}. 
Details regarding dataset statistics\footnote{
We show majority label distributions to highlight common trends while reflecting the diversity of individual annotations.} are shown in Table \ref{tab:distribution}, where the percentage of disagreement is measured by determining the proportion of annotations that differ (by at least one vote) from the majority class (i.e., the hard label) out of the total number of annotations.

\subsubsection{GabHate}

GabHate consists of 27,665 English posts from Gab.com,\footnote{\texttt{\url{https://gab.com}}} each annotated to check if they contain hate speech. Hate speech is defined by \cite{kennedy2018gab} as ``language that dehumanizes, attacks human dignity, derogates, incites violence, or supports hateful ideology, such as white supremacy''. In the dataset, each instance is annotated by at least three annotators from a set of \( \|A\| = 18 \) annotators.

\subsubsection{ConvAbuse} 
The ConvAbuse dataset is the first corpus collecting English dialogues between users and three conversational AI agents, with the goal of detecting varying levels of abusiveness expressed in the text \cite{cercas-curry-etal-2021-convabuse}. The dataset is annotated by multiple annotators, each labeling only a portion of the data. In the original dataset, the annotations ranged from [-3, 1], where 1 indicated no abuse, 0 represented ambivalence, and the remaining values denoted varying degrees of abuse severity. In prior studies \cite{vitsakis2023ilab,leonardelli:semeval}, they consolidated these labels into a binary format, distinguishing between instances where abuse was detected (1) and those where it was not (0). 

\subsubsection{EPIC}
EPIC is the first annotated corpus for irony detection grounded in the principles of data perspectivism \cite{frenda2023epic}. The dataset comprises short social media posts sourced from Twitter\footnote{\texttt{\url{https://twitter.com}}} and Reddit.\footnote{\texttt{\url{https://reddit.com}}} Each post is annotated by multiple annotators with a binary label,   either irony ($1$) or not irony ($0$). 

\subsubsection{StanceDetection}  

The Stance Detection dataset consists of the top 10 news search results retrieved from Google and Bing for 57 queries on various controversial topics. These topics span a diverse range of areas, including education, health, entertainment, religion, and politics, which are known to elicit a variety of perspectives and opinions. 
Each data point consists of a query (about a controversial topic), a document title, and the textual content related to a given query, with varying lengths, i.e., an average of 1,200 tokens, extending up to 7,000 tokens.
Documents (title and content) have been annotated by three annotators on MTurk\footnote{\texttt{\url{https://www.mturk.com}}}. The reported Fleiss-Kappa\footnote{Statistical measure of inter-rater agreement that extends Cohen’s Kappa to multiple raters and accounts for chance agreement.} score of $0.35$ and inter-rater agreement score of $0.49$ on the original dataset highlight the subjectivity and ambiguous nature of the stance detection task. Note that each document has not been annotated by the same three annotators due to the design choices which leads to a more enriched dataset with diverse opinions (annotations).
With the aim of reducing document length and enabling processing with BERT-based models, we applied a summarization step to these documents. For details regarding the summarization method, including model selection, constraints, and performance, please refer to the GitHub repository.

\subsection{Evaluation Metrics}

The model evaluation results are reported using both hard and soft metrics. Hard metrics include accuracy and F1 score (Table \ref{tab:accuracy_f1}), as defined in \cite{uma2021learning}, and commonly used in related work \cite{davani2022dealing}. We opted for macro F1-score to address dataset imbalance and ensure equal importance for each class. These metrics primarily assess model performance by measuring how accurately the model predicts the final (majority) label.
Additionally, we report average model confidence scores (Table \ref{tab:avg_conf}) and Jensen-Shannon Divergence (JSD) (Table \ref{tab:jsd}), following \cite{chen2024seeing}. Confidence scores are calculated by applying the softmax function to the model's logits to obtain class probabilities, then selecting the highest probability as the confidence for each prediction. JSD is used as the principal soft metric to measure the distance between predicted probability distributions (or soft labels).

\section{Results}

\begin{table*}[!t]
    \centering
    \renewcommand{\arraystretch}{0.3}
    \footnotesize
    \begin{tabular}{llccccccccc}
        \toprule
        & & \multicolumn{2}{c}{GabHate} & \multicolumn{2}{c}{ConvAbuse} & \multicolumn{2}{c}{EPIC} & \multicolumn{2}{c}{StanceDetection} \\
        \cmidrule(lr){3-4} \cmidrule(lr){5-6} \cmidrule(lr){7-8} \cmidrule(lr){9-10}
        & \multicolumn{1}{c}{Approach} & \multicolumn{1}{c}{RoBERTa} & \multicolumn{1}{c}{BERT} & \multicolumn{1}{c}{RoBERTa} & \multicolumn{1}{c}{BERT} & \multicolumn{1}{c}{RoBERTa} & \multicolumn{1}{c}{BERT} & \multicolumn{1}{c}{RoBERTa} & \multicolumn{1}{c}{BERT} \\
        \midrule
        \multirow{3}{*}{Accuracy} 
        & Maj. vote & 91.54 & 91.47 & 82.97 & \textbf{82.14} & \textbf{79.11} & 70.88 & 46.76 & 38.84 \\
        & Ensemble & 91.49 & 91.49 & 82.14 & 82.14 & 78.22 & \textbf{77.33} & \textbf{58.99} & \textbf{43.16} \\
        & MultiP & \textbf{91.73} & \textbf{92.21} & \textbf{85.11} & 78.92 & 74.44 & 74.22 & 58.27 & 38.84 \\
        \midrule
        \multirow{3}{*}{Macro-F1} 
        & Maj. vote & 48.63 & 47.77 & \textbf{61.24} & 45.09 & \textit{66.80} & 56.79 & 45.61 & 39.15 \\
        & Ensemble & 47.77 & 47.77 & 45.09 & 45.09 & 59.93 & 47.99 & 59.21 & 43.30 \\
        & MultiP & \textbf{72.26} & \textbf{71.03} & 48.96 & \textbf{57.71} & \textbf{69.38} & \textbf{61.00} & \textbf{61.08} & \textbf{45.22} \\
        \bottomrule
    \end{tabular}
    \caption{Accuracy and Macro-F1 scores across RoBERTa-Large and BERT-Large models.}
    \label{tab:accuracy_f1}
\end{table*}

\begin{table*}[!t]
    \centering
    \renewcommand{\arraystretch}{0.3}
     \footnotesize
    \begin{tabular}{llccccccccc}
        \toprule
        & & \multicolumn{2}{c}{GabHate} & \multicolumn{2}{c}{ConvAbuse} & \multicolumn{2}{c}{EPIC} & \multicolumn{2}{c}{StanceDetection} \\
        \cmidrule(lr){3-4} \cmidrule(lr){5-6} \cmidrule(lr){7-8} \cmidrule(lr){9-10}
        & \multicolumn{1}{c}{Approach} & \multicolumn{1}{c}{RoBERTa} & \multicolumn{1}{c}{BERT} & \multicolumn{1}{c}{RoBERTa} & \multicolumn{1}{c}{BERT} & \multicolumn{1}{c}{RoBERTa} & \multicolumn{1}{c}{BERT} & \multicolumn{1}{c}{RoBERTa} & \multicolumn{1}{c}{BERT} \\
        \midrule
        \multirow{3}{*}{Avg. Conf.} 
        & Maj. vote & 93.40 & 94.84 & 79.73 & 87.74 & \textbf{97.92} & \textbf{93.63} & \textbf{70.48} & \textbf{60.76} \\
        & Ensemble & 95.40 & 94.90 & 86.07 & 87.61 & 83.54 & 79.26 & 47.37 & 50.07 \\
        & MultiP & \textbf{97.55} & \textbf{96.19} & \textbf{98.02} & \textbf{90.84} & 89.35 & 77.61 & 62.60 & 51.18 \\
        \bottomrule
    \end{tabular}
    \caption{Average confidence scores across RoBERTa-Large and BERT-Large models.}
    \label{tab:avg_conf}
\end{table*}

\begin{table*}[!t]
    \centering
    \renewcommand{\arraystretch}{0.3}
        \footnotesize
    \begin{tabular}{llccccccccc}
        \toprule
        & & \multicolumn{2}{c}{GabHate} & \multicolumn{2}{c}{ConvAbuse} & \multicolumn{2}{c}{EPIC} & \multicolumn{2}{c}{StanceDetection} \\
        \cmidrule(lr){3-4} \cmidrule(lr){5-6} \cmidrule(lr){7-8} \cmidrule(lr){9-10}
        & \multicolumn{1}{c}{Approach} & \multicolumn{1}{c}{RoBERTa} & \multicolumn{1}{c}{BERT} & \multicolumn{1}{c}{RoBERTa} & \multicolumn{1}{c}{BERT} & \multicolumn{1}{c}{RoBERTa} & \multicolumn{1}{c}{BERT} & \multicolumn{1}{c}{RoBERTa} & \multicolumn{1}{c}{BERT} \\
        \midrule
        \multirow{3}{*}{JSD} 
        & Maj. vote & 0.388 & 0.694 & 0.138 & 0.245 & 0.655 & 0.548 & 0.281 & 0.297 \\
        & Ensemble & 0.264 & 0.567 & 0.131 & 0.239 & 0.583 & 0.498 & 0.210 & 0.205  \\
        & MultiP & \textbf{0.052} & \textbf{0.051} & \textbf{0.127} & \textbf{0.195} & \textbf{0.134} & \textbf{0.095} & \textbf{0.085} & \textbf{0.062} \\
        \bottomrule
    \end{tabular}
    \caption{Jensen-Shannon Divergence (JSD) scores across RoBERTa-Large and BERT-Large models.}
    \label{tab:jsd}
\end{table*}

The evaluation results demonstrate that multi-perspective models generally outperform baseline approaches— majority vote and ensemble methods—in most subjective tasks (Table \ref{tab:accuracy_f1}). This is particularly evident in hate speech and abusive language detection, where multi-perspective models prove to be more effective at capturing the nuances of the content. Furthermore, both baseline and multi-perspective models tend to perform better in binary classification tasks, such as hate speech, irony, abusive language detection, compared to more complex multi-class tasks like stance detection.
However, one notable exception in F1-score performance is observed with the RoBERTa-large~\cite{liu2019roberta} model fine-tuned on the ConvAbuse dataset. This anomaly can be attributed to the higher agreement among annotators, as indicated by Table \ref{tab:distribution}. In this dataset, the level of disagreement is lower compared to all the other datasets. With greater consensus among annotators, the multi-perspective model faces challenges in capturing diverse viewpoints, making it more difficult to outperform the baseline majority vote. As a result, the majority vote approach achieves a higher F1-score in this case.
An interesting pattern emerges when we examine the confidence levels of the models (Table \ref{tab:avg_conf}). Multi-perspective models tend to exhibit higher confidence in tasks like hate speech and abusive language detection. In contrast, baseline models, especially those using majority vote aggregation, tend to perform better in tasks such as irony and stance detection. This suggests that different models may interpret dissenting opinions and subjective content in distinct ways, with the interpretation depending on the specific task.

The observed uncertainty in multi-perspective models is likely a result of their design, which represents diverse viewpoints with equal importance. While this approach enhances the model's ability to capture dissenting voices, it also increases uncertainty in the predictions. As a result, we argue that confidence scores alone may not reliably evaluate multi-perspective model performance.

To further investigate the performance of multi-perspective models, we use the JSD metric to compare model predictions with human-generated probability distributions, i.e., ground truth soft labels. For baseline models, JSD is computed using softmax predictions, which represent the normalized logits across the classes. Results show that the multi-perspective approach surpasses baselines in all subjective tasks, demonstrating a better ability to align with human label distributions (Table \ref{tab:jsd}).

\section{Ablation Study: XAI}
\label{xai}

To compare model predictions on aggregated (baseline) vs. disaggregated (multi-perspective) labels, we apply  XAI techniques to RoBERTa-large and BERT-large. Using post-hoc\footnote{A post hoc explanation interprets an ML model's predictions after training.} feature-based attribution, we pinpoint the key tokens that influence the model’s decisions and its preference for certain perspectives.

\begin{figure*}[t]
    \centering
    \begin{subfigure}[b]{0.84\textwidth}
        \centering
          \hspace*{-5em}
      \includegraphics[height=3.0em]{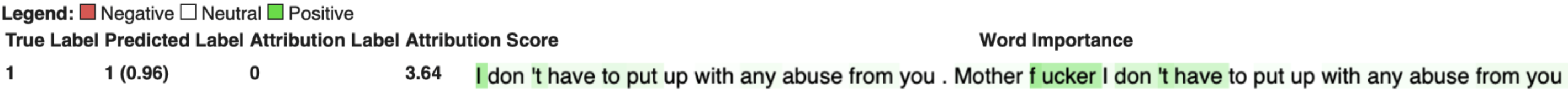}
        \caption{LIG applied as the first post-hoc feature attribution method.}
        \label{fig:conv_less}
    \end{subfigure}
   \begin{subfigure}[b]{0.88\textwidth}
        \centering
         \hspace*{-3em}
        \includegraphics[height=4.6em]{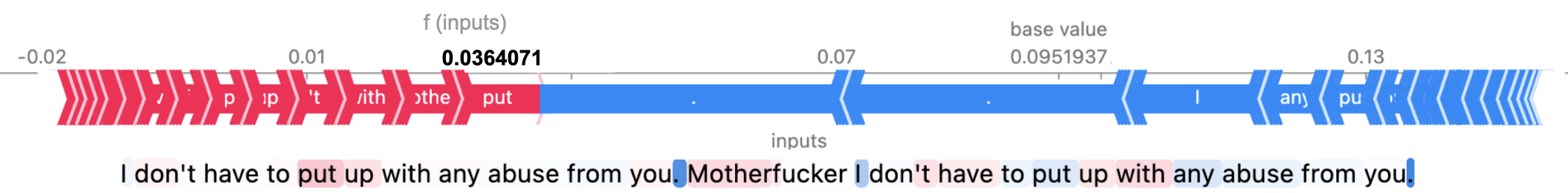}
        \caption{SHAP applied as the second post-hoc feature attribution method.}
        \label{fig:conv_less1}
    \end{subfigure}
    \begin{subfigure}[b]{1.18\textwidth} 
    \centering
     \hspace*{-9em}
    \includegraphics[height=14.0em]{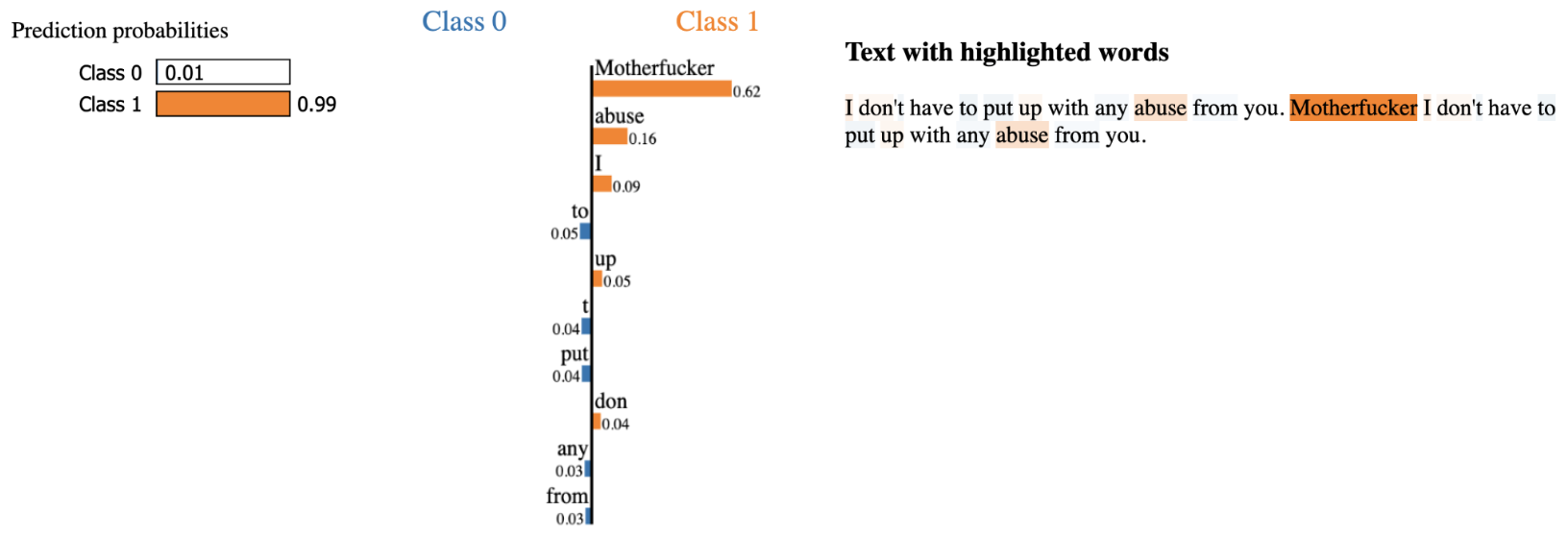}
    \caption{LIME applied as the third post-hoc feature attribution method.}
    \label{fig:conv_less2}
\end{subfigure}
    \caption{Three XAI methods applied to a low-confidence instance identified by the best multi-perspective model on ConvAbuse.}
    \label{fig:conv_explanations}
\end{figure*}

\subsection{Feature-based Methods}
There is growing interest in the NLP community in better understanding the uncertainty associated with LLM outputs, which are often regarded as black boxes as their interpretability remains challenging~\cite{wu2024usable,ahdritz:distinguishing}. To gain insights into model predictions, we explore various XAI techniques, in particular:  
\begin{enumerate}
    \item Layer Integrated Gradient (LIG) \cite{sundararajan2017axiomatic} is a variant of the Integrated Gradient algorithm that computes importance scores for input features by approximating the integral of the model's output across different layers along a baseline path\footnote{It is a zero vector representing no feature information.}. 
    \item LIME \cite{ribeiro2016should} explains model predictions by auditing a black-box model on synthetic instances near the target instance, estimating feature importance using a locally fitted linear regression. Higher coefficients indicate greater influence, providing insights into local decision boundaries and model reasoning.
\item SHAP \cite{lundberg2017unified}, similarly to LIME, SHAP provides feature importance scores leveraging Shapley values to calculate additive feature attributions. High values indicate a strong positive contribution to the classification outcome, whereas values near zero or above suggest a negative or negligible impact.

\item Layer Conductance (LC) \cite{dhamdhere2018important}, this method evaluates the conductance of each hidden layer, identifying the neurons that contributed the most to the model's prediction within a target layer.
\item Attention scores indicate how much focus each token gives to others in the sequence across layers and attention heads. They provide a direct visualization of the model’s focus distribution, revealing key relationships that influence its decision-making.
\end{enumerate}

All the aforementioned methods, except for attention matrices, are implemented using Captum\footnote{\texttt{\url{https://captum.ai}}} which is an open-source model interpretability library for Pytorch~\cite{kokhlikyan2020captum}.

\begin{figure*}[t]  
    \centering
    \begin{subfigure}[b]{0.84\textwidth}
        \centering
         \hspace*{-5em}
          \includegraphics[height=4.3em]
         {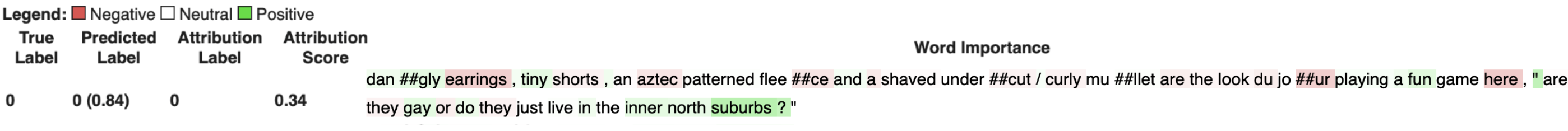}
        \caption{LIG applied as the first post-hoc feature attribution method.}
        \label{fig:lig_epic1} 
    \end{subfigure}
     \begin{subfigure}[b]{0.90\textwidth}
        \centering
        \hspace*{-2em}
        \includegraphics[height=5.5em]{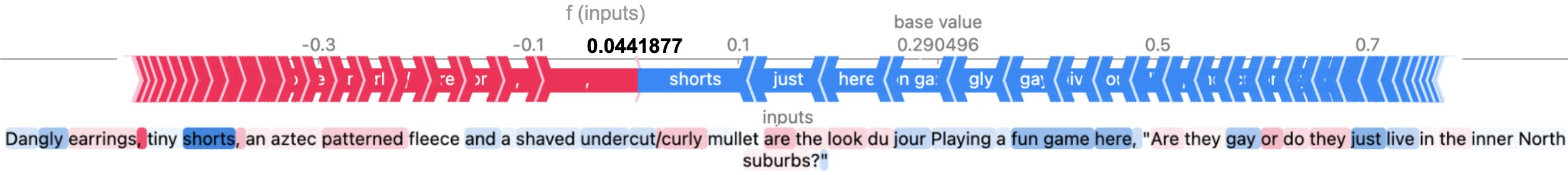}
        \caption{SHAP applied as the second post-hoc feature attribution method.}
        \label{fig:lig_epic2}
    \end{subfigure}
    \hfill
     \begin{subfigure}[b]{1.18\textwidth}
    \centering
     \hspace*{-5em}
    \includegraphics[height=14.0em]{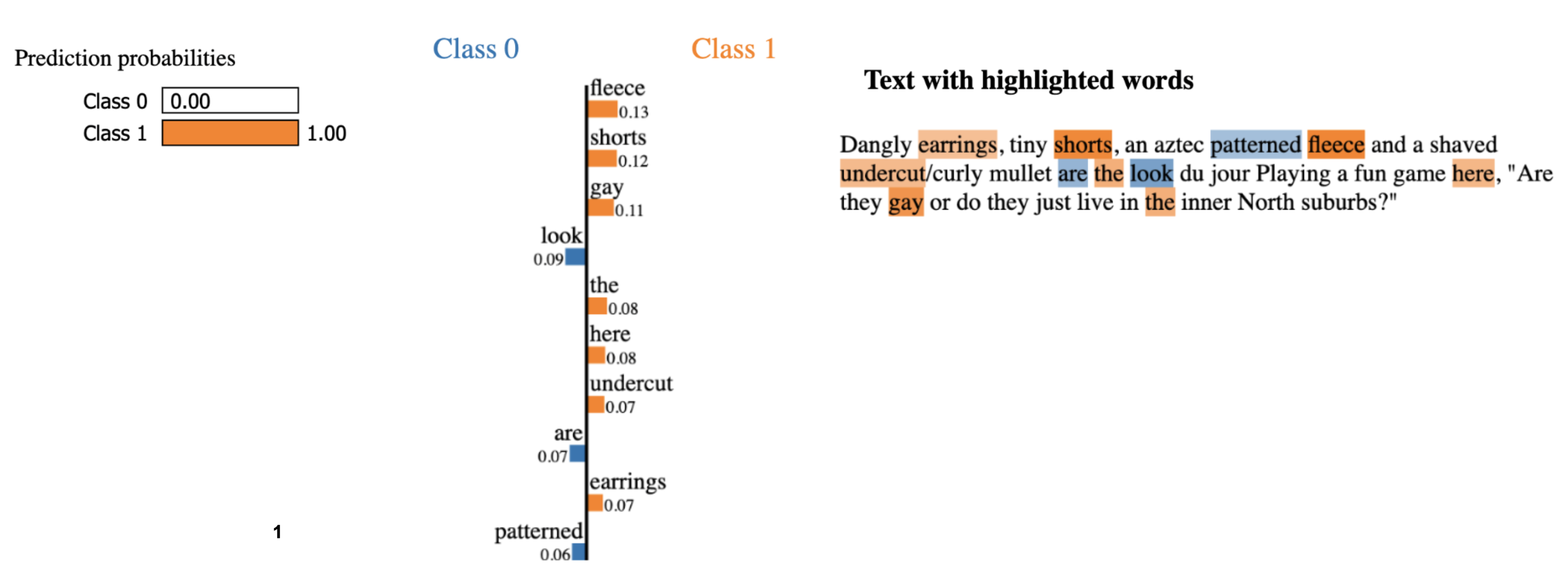}
        \caption{LIME applied as the third post-hoc feature attribution method.}
        \label{fig:lig_epic3} 
    \end{subfigure}
     \caption{Three XAI methods applied to a low-confidence instance identified by the best baseline model on EPIC.}
    \label{fig:epic_explanations}  
\end{figure*}

\subsection{Results} 
\label{sec:results_xai}
We apply the aforementioned XAI methods to the top-performing models: the baseline majority vote for the EPIC dataset and the multi-perspective model for ConvAbuse. Model selection was based on average confidence scores on the respective test sets. We compute the prediction confidence (i.e., probability) for each test instance and ranked the instances accordingly.
We then select the five instances with the lowest prediction probabilities (i.e., highest uncertainty) and, due to space constraints, present the two most representative examples.
A key factor in feature-based attribution is the number of salient tokens (\textit{k}). Following \cite{krishna2022disagreement}, we set \textit{k} based on average sentence length to ensure a balanced meaningful token selection.
Some interesting insights emerge from the models' lower-confidence predictions. Figure \ref{fig:conv_explanations} illustrates the application of LIG, SHAP, and LIME on ConvAbuse dataset on the task of Abusive Language Detection. 
The example shown is classified as abusive language (1) by the best-performing multi-perspective model (black box) with a confidence score of $0.55$. The true (hard) label for this instance is 1, while its soft label representation is [0.0, 1.0]\footnote{At index 0 class 0, at index 1 class 1.}, meaning all annotators agreed on label 1, with no perspectives discarded.
In Figure \ref{fig:conv_less}, LIG highlights the key contributing words in green—‘fucker’ as the strongest (darkest shade), followed by ‘don’t’ and ‘have’ (lighter green).
Figure~\ref{fig:conv_less1} shows a force plot with SHAP values for each word contributing to the model’s output of 0.036 (i.e. class 0, non-abusive). Red indicates features increasing the prediction probability, while blue indicates features decreasing it.
Figure~\ref{fig:conv_less2} shows that the tokens `motherfucker' and `abuse' contribute the most to the prediction probability, with weights of 0.63 and 0.15 respectively, followed by `I' and `up' with lower contributions of 0.09 and 0.06, demonstrating LIME’s accurate identification of key influential features.
Among the three XAI methods, LIG and LIME consistently predict class 1. However, each method produces slightly different explanations, highlighting variability in token importance depending on the perspective.
Figure~\ref{fig:epic_explanations} analyzes a low-confidence instance (model confidence score of 0.57) from the EPIC dataset (Irony Detection), classified by the best-performing baseline model (black box).
The true (hard) label is 0 (non-ironic content), while the soft label representation [0.9, 0.1] indicates slight disagreement.
In this case, LIG classifies the instance as class 0, emphasizing tokens such as `tiny', `fun', `gay', and `suburbs', with suburbs having the strongest positive influence, indicated by a dark green shade.
Similarly, SHAP assigns a prediction value of 0.044 (corresponding to class 0, non-ironic), while LIME confidently predicts class 1.
Figure \ref{fig:lig_epic2} shows that the tokens of `shorts', and `gay' (highlighted in blue) reduce the prediction probability.
In contrast, Figure~\ref{fig:lig_epic3} indicates that the tokens `fleece', `shorts', and `gay' contributed most to the incorrect prediction of class 1 (ironic content), with weights of 0.14, 0.11, and 0.09, respectively.
Unlike the abusive language detection task, LIG and SHAP correctly agree on the prediction (class 0) for irony detection. However, they highlight different tokens, illustrating variability in token importance across methods. This underscores the limitations of relying on a single explainability technique. Additional analyses of token importance across layers and attention patterns are available in our GitHub repository, linked on the first page.

\section{Conclusion \& Future Work}

This work presents a novel multi-perspective approach for subjective tasks. We advocate adopting perspective-aware models that preserve, rather than aggregate, diverse human opinions to promote responsible AI.
%
The perspectivist approach promotes inclusivity and critically addresses assumptions inherent in the use of LLMs within multimodal contexts involving image, audio, and video analysis. Furthermore, it has potential applications in traditionally objective tasks, such as medical decision-making \cite{basile2021toward}.
%
In this study, we extended prior work by introducing a methodology based on soft labels to avoid overshadowing minority perspectives, particularly in the absence of a definitive ground truth. We performed an extensive analysis across diverse subjective tasks—including hate speech, irony, and stance detection—to capture and preserve multiple viewpoints.
%
We fine-tuned BERT-large and RoBERTa-large models using both a baseline method—majority vote with ensemble—and a multi-perspective approach. Results indicate that employing soft labels enhances model performance (e.g., F1 score) and more accurately reflects human label distributions \cite{chen2024seeing}. Additionally, we applied explainable AI techniques to identify key tokens influencing model predictions.
%
Our findings expose inconsistencies in post-hoc attribution methods (LIG, SHAP, LIME), underscoring the inherent subjectivity of target tasks, where interpretations vary according to annotators’ backgrounds, experiences, and values. This aligns with the variability observed in human judgments. Further investigation into the correlation between attribution variability and human subjectivity is warranted. We advocate adopting a pluralistic approach \cite{sorensen2024roadmap} to better align LLMs with diverse human values, as the majority viewpoint may not always be the most appropriate.
This paradigm has the potential to lead to a new generation of more inclusive LLMs that avoid falling into the echo chamber loop \cite{nehring2024large}, bringing attention to even the minority perspectives that are often obscured. Thus, from a practical perspective, this approach could enhance the trustworthiness of LLMs and support the development of more personalized systems \cite{huang2024position}, ultimately improving multi-turn human–AI interactions, particularly in chatbot applications.

In future work, we plan to extend our approach to a broader range of models, including decoder-only architectures prompted for classification in both subjective and objective tasks. Additionally, we will investigate the relationship between model uncertainty and human disagreement. Moreover, we aim to employ advanced XAI techniques, such as neuron activation analysis, to enhance interpretability and facilitate bias detection.

%

\section{Limitations}
%
This study has certain limitations. While excluding instances without a majority label reduced the dataset size, this was essential to maintain a fair comparison with baseline models. Nevertheless, these instances offer important insights into diverse perspectives shaped by individuals’ distinct backgrounds and experiences, and should not be disregarded as noise.
In future work, we intend to incorporate high-disagreement instances and perform a more in-depth analysis of their effects. Our study was also limited by computational resources, which impacted batch size and model capacity. Finally, current XAI methods in NLP frequently fail to deliver the clarity of insights observed in other domains.


\section*{Acknowledgments}
This work has been supported by the European Union under ERC-2018-ADG
GA 834756 (XAI), the Partnership Extended PE00000013 - “FAIR - Future Artificial Intelligence Research” - Spoke 1 “Human-centered AI”,  and by the EU EIC project EMERGE (Grant No. 101070918).  

\newpage

\begin{thebibliography}{}

\bibitem[\protect\citeauthoryear{Abercrombie \bgroup \em et al.\egroup }{2023}]{abercrombie:consistency}
Gavin Abercrombie, Verena Rieser, and Dirk Hovy.
\newblock Consistency is key: Disentangling label variation in natural language processing with intra-annotator agreement.
\newblock {\em arXiv preprint arXiv:2301.10684}, 2023.

\bibitem[\protect\citeauthoryear{Ahdritz \bgroup \em et al.\egroup }{2024}]{ahdritz:distinguishing}
Gustaf Ahdritz, Tian Qin, Nikhil Vyas, Boaz Barak, and Benjamin~L Edelman.
\newblock Distinguishing the knowable from the unknowable with language models.
\newblock {\em arXiv preprint arXiv:2402.03563}, 2024.

\bibitem[\protect\citeauthoryear{Aoyagui \bgroup \em et al.\egroup }{2024}]{aoyagui2024exploring}
Paula~Akemi Aoyagui, Sharon Ferguson, and Anastasia Kuzminykh.
\newblock Exploring subjectivity for more human-centric assessment of social biases in large language models.
\newblock {\em arXiv preprint arXiv:2405.11048}, 2024.

\bibitem[\protect\citeauthoryear{Basile and others}{2020}]{basile:2020s}
Valerio Basile et~al.
\newblock It’s the end of the gold standard as we know it. on the impact of pre-aggregation on the evaluation of highly subjective tasks.
\newblock In {\em CEUR workshop proceedings}, volume 2776, pages 31--40. CEUR-WS, 2020.

\bibitem[\protect\citeauthoryear{Basile \bgroup \em et al.\egroup }{2021}]{basile2021toward}
Valerio Basile, Federico Cabitza, Andrea Campagner, and Michael Fell.
\newblock Toward a perspectivist turn in ground truthing for predictive computing.
\newblock {\em arXiv preprint arXiv:2109.04270}, 2021.

\bibitem[\protect\citeauthoryear{Beck \bgroup \em et al.\egroup }{2024}]{beck2024sensitivity}
Tilman Beck, Hendrik Schuff, Anne Lauscher, and Iryna Gurevych.
\newblock Sensitivity, performance, robustness: Deconstructing the effect of sociodemographic prompting.
\newblock In {\em Proceedings of the 18th Conference of the European Chapter of the Association for Computational Linguistics (Volume 1: Long Papers)}, pages 2589--2615, 2024.

\bibitem[\protect\citeauthoryear{Cercas~Curry \bgroup \em et al.\egroup }{2021}]{cercas-curry-etal-2021-convabuse}
Amanda Cercas~Curry, Gavin Abercrombie, and Verena Rieser.
\newblock {C}onv{A}buse: Data, analysis, and benchmarks for nuanced abuse detection in conversational {AI}.
\newblock In Marie-Francine Moens, Xuanjing Huang, Lucia Specia, and Scott Wen-tau Yih, editors, {\em Proceedings of the 2021 Conference on Empirical Methods in Natural Language Processing}, pages 7388--7403, Online and Punta Cana, Dominican Republic, November 2021. Association for Computational Linguistics.

\bibitem[\protect\citeauthoryear{Chen \bgroup \em et al.\egroup }{2024}]{chen2024seeing}
Beiduo Chen, Xinpeng Wang, Siyao Peng, Robert Litschko, Anna Korhonen, and Barbara Plank.
\newblock “seeing the big through the small”: Can llms approximate human judgment distributions on nli from a few explanations?
\newblock In {\em Findings of the Association for Computational Linguistics: EMNLP 2024}, pages 14396--14419, 2024.

\bibitem[\protect\citeauthoryear{Collins \bgroup \em et al.\egroup }{2022}]{collins2022eliciting}
Katherine~M Collins, Umang Bhatt, and Adrian Weller.
\newblock Eliciting and learning with soft labels from every annotator.
\newblock In {\em Proceedings of the AAAI conference on human computation and crowdsourcing}, volume~10, pages 40--52, 2022.

\bibitem[\protect\citeauthoryear{Devlin}{2018}]{devlin2018bert}
Jacob Devlin.
\newblock Bert: Pre-training of deep bidirectional transformers for language understanding.
\newblock {\em arXiv preprint arXiv:1810.04805}, 2018.

\bibitem[\protect\citeauthoryear{Dhamdhere \bgroup \em et al.\egroup }{2018}]{dhamdhere2018important}
Kedar Dhamdhere, Mukund Sundararajan, and Qiqi Yan.
\newblock How important is a neuron?
\newblock {\em arXiv preprint arXiv:1805.12233}, 2018.

\bibitem[\protect\citeauthoryear{Fleisig \bgroup \em et al.\egroup }{2023}]{fleisig:majority}
Eve Fleisig, Rediet Abebe, and Dan Klein.
\newblock When the majority is wrong: Modeling annotator disagreement for subjective tasks.
\newblock In {\em Proceedings of the 2023 Conference on Empirical Methods in Natural Language Processing}, pages 6715--6726, 2023.

\bibitem[\protect\citeauthoryear{Frenda \bgroup \em et al.\egroup }{2023}]{frenda2023epic}
Simona Frenda, Alessandro Pedrani, Valerio Basile, Soda~Marem Lo, Alessandra~Teresa Cignarella, Raffaella Panizzon, Cristina S{\'a}nchez-Marco, Bianca Scarlini, Viviana Patti, Cristina Bosco, et~al.
\newblock Epic: multi-perspective annotation of a corpus of irony.
\newblock In {\em Proceedings of the 61st Annual Meeting of the Association for Computational Linguistics (Volume 1: Long Papers)}, pages 13844--13857, 2023.

\bibitem[\protect\citeauthoryear{Frenda \bgroup \em et al.\egroup }{2024}]{frenda2024perspectivist}
Simona Frenda, Gavin Abercrombie, Valerio Basile, Alessandro Pedrani, Raffaella Panizzon, Alessandra~Teresa Cignarella, Cristina Marco, and Davide Bernardi.
\newblock Perspectivist approaches to natural language processing: a survey.
\newblock {\em Language Resources and Evaluation}, pages 1--28, 2024.

\bibitem[\protect\citeauthoryear{Gezici \bgroup \em et al.\egroup }{2021}]{gezici2021evaluation}
Gizem Gezici, Aldo Lipani, Yucel Saygin, and Emine Yilmaz.
\newblock Evaluation metrics for measuring bias in search engine results.
\newblock {\em Information Retrieval Journal}, 24:85--113, 2021.

\bibitem[\protect\citeauthoryear{Hinton \bgroup \em et al.\egroup }{2015}]{hinton2015distilling}
Geoffrey Hinton, Oriol Vinyals, and Jeff Dean.
\newblock Distilling the knowledge in a neural network.
\newblock {\em arXiv preprint arXiv:1503.02531}, 2015.

\bibitem[\protect\citeauthoryear{Huang \bgroup \em et al.\egroup }{2024}]{huang2024position}
Yue Huang, Lichao Sun, Haoran Wang, Siyuan Wu, Qihui Zhang, Yuan Li, Chujie Gao, Yixin Huang, Wenhan Lyu, Yixuan Zhang, et~al.
\newblock Position: Trustllm: Trustworthiness in large language models.
\newblock In {\em International Conference on Machine Learning}, pages 20166--20270. PMLR, 2024.

\bibitem[\protect\citeauthoryear{Kennedy \bgroup \em et al.\egroup }{2018}]{kennedy2018gab}
Brendan Kennedy, Mohammad Atari, Aida~Mostafazadeh Davani, Leigh Yeh, Ali Omrani, Yehsong Kim, Kris Coombs, Shreya Havaldar, Gwenyth Portillo-Wightman, Elaine Gonzalez, et~al.
\newblock The gab hate corpus: A collection of 27k posts annotated for hate speech.
\newblock {\em PsyArXiv. July}, 18, 2018.

\bibitem[\protect\citeauthoryear{Kirk \bgroup \em et al.\egroup }{2024}]{kirk:benefits}
Hannah~Rose Kirk, Bertie Vidgen, Paul R{\"o}ttger, and Scott~A Hale.
\newblock The benefits, risks and bounds of personalizing the alignment of large language models to individuals.
\newblock {\em Nature Machine Intelligence}, pages 1--10, 2024.

\bibitem[\protect\citeauthoryear{Kokhlikyan \bgroup \em et al.\egroup }{2020}]{kokhlikyan2020captum}
Narine Kokhlikyan, Vivek Miglani, Miguel Martin, Edward Wang, Bilal Alsallakh, Jonathan Reynolds, Alexander Melnikov, Natalia Kliushkina, Carlos Araya, Siqi Yan, et~al.
\newblock Captum: A unified and generic model interpretability library for pytorch.
\newblock {\em arXiv preprint arXiv:2009.07896}, 2020.

\bibitem[\protect\citeauthoryear{Krishna \bgroup \em et al.\egroup }{2022}]{krishna2022disagreement}
Satyapriya Krishna, Tessa Han, Alex Gu, Steven Wu, Shahin Jabbari, and Himabindu Lakkaraju.
\newblock The disagreement problem in explainable machine learning: A practitioner's perspective.
\newblock {\em arXiv preprint arXiv:2202.01602}, 2022.

\bibitem[\protect\citeauthoryear{Leonardelli \bgroup \em et al.\egroup }{2023}]{leonardelli:semeval}
Elisa Leonardelli, Alexandra Uma, Gavin Abercrombie, Dina Almanea, Valerio Basile, Tommaso Fornaciari, Barbara Plank, Verena Rieser, and Massimo Poesio.
\newblock Semeval-2023 task 11: Learning with disagreements (lewidi).
\newblock {\em arXiv preprint arXiv:2304.14803}, 2023.

\bibitem[\protect\citeauthoryear{Liu}{2019}]{liu2019roberta}
Y~Liu.
\newblock Roberta: A robustly optimized bert pretraining approach.
\newblock {\em arXiv preprint arXiv:1907.11692}, 2019.

\bibitem[\protect\citeauthoryear{Lundberg}{2017}]{lundberg2017unified}
Scott Lundberg.
\newblock A unified approach to interpreting model predictions.
\newblock {\em arXiv preprint arXiv:1705.07874}, 2017.

\bibitem[\protect\citeauthoryear{Mostafazadeh~Davani \bgroup \em et al.\egroup }{2022}]{davani2022dealing}
Aida Mostafazadeh~Davani, Mark D{\'i}az, and Vinodkumar Prabhakaran.
\newblock Dealing with disagreements: Looking beyond the majority vote in subjective annotations.
\newblock {\em Transactions of the Association for Computational Linguistics}, 10:92--110, 2022.

\bibitem[\protect\citeauthoryear{Muscato \bgroup \em et al.\egroup }{2024}]{muscato2024multi}
Benedetta Muscato, Praveen Bushipaka, Gizem Gezici, Lucia Passaro, Fosca Giannotti, et~al.
\newblock Multi-perspective stance detection.
\newblock In {\em CEUR WORKSHOP PROCEEDINGS}, volume 3825, pages 208--214. CEUR-WS, 2024.

\bibitem[\protect\citeauthoryear{Muscato \bgroup \em et al.\egroup }{2025}]{muscato2025embracing}
Benedetta Muscato, Praveen Bushipaka, Gizem Gezici, Lucia Passaro, Fosca Giannotti, and Tommaso Cucinotta.
\newblock Embracing diversity: A multi-perspective approach with soft labels.
\newblock {\em arXiv preprint arXiv:2503.00489}, 2025.

\bibitem[\protect\citeauthoryear{Nehring \bgroup \em et al.\egroup }{2024}]{nehring2024large}
Jan Nehring, Aleksandra Gabryszak, Pascal J{\"u}rgens, Aljoscha Burchardt, Stefan Schaffer, Matthias Spielkamp, and Birgit Stark.
\newblock Large language models are echo chambers.
\newblock In {\em Proceedings of the 2024 Joint International Conference on Computational Linguistics, Language Resources and Evaluation (LREC-COLING 2024)}, pages 10117--10123, 2024.

\bibitem[\protect\citeauthoryear{Pereyra \bgroup \em et al.\egroup }{2017}]{pereyra:regularizing}
Gabriel Pereyra, George Tucker, Jan Chorowski, {\L}ukasz Kaiser, and Geoffrey Hinton.
\newblock Regularizing neural networks by penalizing confident output distributions.
\newblock {\em arXiv preprint arXiv:1701.06548}, 2017.

\bibitem[\protect\citeauthoryear{Plank}{2022}]{plank2022problem}
Barbara Plank.
\newblock The'problem'of human label variation: On ground truth in data, modeling and evaluation.
\newblock {\em arXiv preprint arXiv:2211.02570}, 2022.

\bibitem[\protect\citeauthoryear{Ribeiro \bgroup \em et al.\egroup }{2016}]{ribeiro2016should}
Marco~Tulio Ribeiro, Sameer Singh, and Carlos Guestrin.
\newblock " why should i trust you?" explaining the predictions of any classifier.
\newblock In {\em Proceedings of the 22nd ACM SIGKDD international conference on knowledge discovery and data mining}, pages 1135--1144, 2016.

\bibitem[\protect\citeauthoryear{Sandri \bgroup \em et al.\egroup }{2023}]{sandri:don}
Marta Sandri, Elisa Leonardelli, Sara Tonelli, and Elisabetta Je{\v{z}}ek.
\newblock Why don’t you do it right? analysing annotators’ disagreement in subjective tasks.
\newblock In {\em Proceedings of the 17th Conference of the European Chapter of the Association for Computational Linguistics}, pages 2428--2441, 2023.

\bibitem[\protect\citeauthoryear{Santurkar \bgroup \em et al.\egroup }{2023}]{santurkar2023whose}
Shibani Santurkar, Esin Durmus, Faisal Ladhak, Cinoo Lee, Percy Liang, and Tatsunori Hashimoto.
\newblock Whose opinions do language models reflect?
\newblock In {\em International Conference on Machine Learning}, pages 29971--30004. PMLR, 2023.

\bibitem[\protect\citeauthoryear{Schmeisser-Nieto \bgroup \em et al.\egroup }{2024}]{schmeisser2024human}
Wolfgang~S Schmeisser-Nieto, Pol Pastells, Simona Frenda, and Mariona Taul{\'e}.
\newblock Human vs. machine perceptions on immigration stereotypes.
\newblock In {\em Proceedings of the 2024 Joint International Conference on Computational Linguistics, Language Resources and Evaluation (LREC-COLING 2024)}, pages 8453--8463, 2024.

\bibitem[\protect\citeauthoryear{Sorensen \bgroup \em et al.\egroup }{2024}]{sorensen2024roadmap}
Taylor Sorensen, Jared Moore, Jillian Fisher, Mitchell Gordon, Niloofar Mireshghallah, Christopher~Michael Rytting, Andre Ye, Liwei Jiang, Ximing Lu, Nouha Dziri, et~al.
\newblock A roadmap to pluralistic alignment.
\newblock {\em arXiv preprint arXiv:2402.05070}, 2024.

\bibitem[\protect\citeauthoryear{Sundararajan \bgroup \em et al.\egroup }{2017}]{sundararajan2017axiomatic}
Mukund Sundararajan, Ankur Taly, and Qiqi Yan.
\newblock Axiomatic attribution for deep networks.
\newblock In {\em International conference on machine learning}, pages 3319--3328. PMLR, 2017.

\bibitem[\protect\citeauthoryear{Szegedy \bgroup \em et al.\egroup }{2016}]{szegedy2016rethinking}
Christian Szegedy, Vincent Vanhoucke, Sergey Ioffe, Jon Shlens, and Zbigniew Wojna.
\newblock Rethinking the inception architecture for computer vision.
\newblock In {\em Proceedings of the IEEE conference on computer vision and pattern recognition}, pages 2818--2826, 2016.

\bibitem[\protect\citeauthoryear{Uma \bgroup \em et al.\egroup }{2020}]{uma2020case}
Alexandra Uma, Tommaso Fornaciari, Dirk Hovy, Silviu Paun, Barbara Plank, and Massimo Poesio.
\newblock A case for soft loss functions.
\newblock In {\em Proceedings of the AAAI Conference on Human Computation and Crowdsourcing}, volume~8, pages 173--177, 2020.

\bibitem[\protect\citeauthoryear{Uma \bgroup \em et al.\egroup }{2021}]{uma2021learning}
Alexandra~N Uma, Tommaso Fornaciari, Dirk Hovy, Silviu Paun, Barbara Plank, and Massimo Poesio.
\newblock Learning from disagreement: A survey.
\newblock {\em Journal of Artificial Intelligence Research}, 72:1385--1470, 2021.

\bibitem[\protect\citeauthoryear{Vitsakis \bgroup \em et al.\egroup }{2023}]{vitsakis2023ilab}
Nikolas Vitsakis, Amit Parekh, Tanvi Dinkar, Gavin Abercrombie, Ioannis Konstas, and Verena Rieser.
\newblock ilab at semeval-2023 task 11 le-wi-di: Modelling disagreement or modelling perspectives?
\newblock {\em arXiv preprint arXiv:2305.06074}, 2023.

\bibitem[\protect\citeauthoryear{Wang \bgroup \em et al.\egroup }{2023}]{wang:aligning}
Yufei Wang, Wanjun Zhong, Liangyou Li, Fei Mi, Xingshan Zeng, Wenyong Huang, Lifeng Shang, Xin Jiang, and Qun Liu.
\newblock Aligning large language models with human: A survey.
\newblock {\em arXiv preprint arXiv:2307.12966}, 2023.

\bibitem[\protect\citeauthoryear{Wu \bgroup \em et al.\egroup }{2024}]{wu2024usable}
Xuansheng Wu, Haiyan Zhao, Yaochen Zhu, Yucheng Shi, Fan Yang, Tianming Liu, Xiaoming Zhai, Wenlin Yao, Jundong Li, Mengnan Du, et~al.
\newblock Usable xai: 10 strategies towards exploiting explainability in the llm era.
\newblock {\em arXiv preprint arXiv:2403.08946}, 2024.

\end{thebibliography}

\end{document}